\documentclass[11pt,letterpaper]{article}
\pdfoutput=1
\usepackage{naaclhlt2016}
\usepackage{times}
\usepackage{latexsym}
\usepackage{amsmath}
\usepackage{multirow}
\usepackage{url}
\usepackage{graphicx}
\usepackage{amssymb}
\usepackage{amsfonts}
\usepackage[hidelinks]{hyperref}
\usepackage{fancyvrb}
\newcommand{\citep}[1]{\cite{#1}}
\newcommand{\citet}[1]{\newcite{#1}}

\naaclfinalcopy 
\def\naaclpaperid{132} 

\usepackage{xcolor}
\newcommand{\CHANGEA}[1]{#1}   
\newcommand{\CHANGEB}[1]{#1}   

\title{STransE: a novel embedding model of entities and relationships \\ in knowledge bases\thanks{\ \ A revised version of our NAACL-HLT 2016 paper with additional experimental results and latest related work.}}

\author{Dat Quoc Nguyen${}^{1}$, Kairit Sirts${}^{1}$, Lizhen Qu${}^{2}$ \and Mark Johnson${}^{1}$ \\
\\
${}^{1}$ Department of Computing, Macquarie University, Sydney, Australia \\
{\tt{\small{dat.nguyen@students.mq.edu.au, \{kairit.sirts, mark.johnson\}@mq.edu.au}}} \\
${}^{2}$ NICTA, ACT 2601, Australia \\
{\tt{\small{lizhen.qu@nicta.com.au}}}
}

\begin{document}

\maketitle

\begin{abstract}
Knowledge bases of real-world facts about entities and their relationships are useful resources for a variety of natural language processing tasks. However, because knowledge bases are typically incomplete, it is useful to be able to perform \textit{link prediction} or \emph{knowledge base completion}, i.e., predict whether a relationship not in the knowledge base is likely to be true.  \CHANGEA{This paper combines insights \CHANGEB{from} several previous link prediction models into a new embedding model \textit{STransE}} that represents each entity as a low-dimensional vector, and each relation by two matrices and a translation vector. \CHANGEA{STransE is \CHANGEB{a simple combination of the SE and TransE  models,} but} \CHANGEA{it} obtains better link prediction performance on two benchmark datasets than previous embedding models. \CHANGEA{Thus, \CHANGEB{STransE can serve as} a new baseline for the more complex models in \CHANGEB{the} link prediction task.}
\end{abstract}

\section{Introduction} \label{sec:intro}
Knowledge bases (KBs), such as WordNet \cite{FellbaumC98}, YAGO
\cite{Suchanek:2007}, Freebase \cite{Bollacker:2008} and DBpedia
\cite{LehmannIJJKMHMK15}, represent relationships between entities as
triples $(\mathrm{head\ entity, relation, tail\ entity})$.  Even very
large knowledge bases are still far from complete
\cite{NIPS2013_5028,West:2014:KBC:2566486.2568032}.
\emph{Link prediction} or \emph{knowledge base completion} systems
\cite{NickelMTG15} predict which triples not in a knowledge base are
likely to be true  \cite{NIPS2003_2465,bordes-2011}.  A variety of
different kinds of information is potentially useful here, including
information extracted from external corpora
\cite{riedel-EtAl:2013:NAACL-HLT,wang-EtAl:2014:EMNLP20145} and the
other relationships that hold between the entities
\cite{angeli-manning:2013:CoNLL-2013,zhao2015y}.
\CHANGEB{For example, \newcite{toutanova-EtAl:2015:EMNLP} used information
from the external ClueWeb-12 corpus to significantly enhance performance.}

While integrating a wide variety of information sources can produce excellent results \citep{DasNBM17},
there are several reasons for studying simpler models that
directly optimize a score function for the triples in a knowledge base, such as the one presented here.
\CHANGEB{First, additional information sources might not be available, e.g., for knowledge bases for specialized domains.}
\CHANGEB{Second}, models \CHANGEB{that don't exploit external resources} are simpler and thus typically much faster to train than the more complex models using additional information. 
\CHANGEB{Third}, the more complex models \CHANGEB{that exploit external information} 
are typically extensions of these simpler models,
and are often initialized with parameters estimated by such simpler models,
so improvements to the simpler models should yield corresponding improvements
to the more complex models \CHANGEB{as well}.

\begin{table*}[ht]
\centering
{\footnotesize
\resizebox{16.5cm}{!}{
\def\arraystretch{1.25}
\begin{tabular}{l|l|l}
\hline
\bf Model & Score function $f_r(h, t)$ & \bf Opt.  \\
\hline
\hline
SE &  $\| \textbf{W}_{r,1}\textbf{h} - \textbf{W}_{r,2}\textbf{t}\|_{\ell_{1/2}}$ ;  $\textbf{W}_{r,1}$, $\textbf{W}_{r,2}$ $\in$ $\mathbb{R}^{k \times k}$ & SGD\\
\hline
Unstructured &  $ \| \textbf{h} - \textbf{t} \|_{\ell_{1/2}}$ & SGD \\
\hline
TransE &  $ \|  \textbf{h} + \textbf{r} - \textbf{t} \|_{\ell_{1/2}}$ ;    \textbf{r} $\in$ $\mathbb{R}^{k}$ & SGD \\
\hline
DISTMULT &  $ \textbf{h}^{\top}\textbf{W}_{r}\textbf{t}$ ;   $\textbf{W}_{r}$ is a diagonal matrix $\in$ $\mathbb{R}^{k \times k}$ & AdaGrad\\
\hline
{NTN} & $ \textbf{u}_r^{\top} \mathit{tanh}( \textbf{h}^{\top}\textbf{M}_{r}\textbf{t}  + \textbf{W}_{r,1}\textbf{h} + \textbf{W}_{r,2}\textbf{t} + \textbf{b}_r)$ ; $\textbf{u}_r\text{, } \textbf{b}_r \in \mathbb{R}^d$; $\textbf{M}_{r} \in \mathbb{R}^{k \times k \times d}$;  $\textbf{W}_{r,1}$, $\textbf{W}_{r,2} \in \mathbb{R}^{d \times k}$ & {L-BFGS}  \\
\hline
{TransH} & $\| (\textbf{I} - \textbf{r}_p\textbf{r}_p^{\top})\textbf{h} + \textbf{r} - (\textbf{I} - \textbf{r}_p\textbf{r}_p^{\top})\textbf{t} \|_{\ell_{1/2}}$ ; $\textbf{r}_p$, $\textbf{r} \in$ $\mathbb{R}^{k}$ ; $\textbf{I}$: Identity matrix size $k \times k$ & {SGD}\\
\hline
{TransD} & $\| (\textbf{I} + \textbf{r}_p\textbf{h}_p^{\top})\textbf{h} + \textbf{r} - (\textbf{I} + \textbf{r}_p\textbf{t}_p^{\top})\textbf{t} \|_{\ell_{1/2}}$ ;  \textbf{r}$_p$, \textbf{r} $\in$ $\mathbb{R}^{d}$ ; \textbf{h}$_p$, \textbf{t}$_p$ $\in$ $\mathbb{R}^{k}$ ; $\textbf{I}$: Identity matrix size $d \times k$  & {AdaDelta} \\
\hline
TransR & $\| \textbf{W}_{r}\textbf{h} + \textbf{r} - \textbf{W}_{r}\textbf{t}\|_{\ell_{1/2}}$ ;  $\textbf{W}_{r}$ $\in$ $\mathbb{R}^{d \times k}$  ;    \textbf{r} $\in$ $\mathbb{R}^{d}$ & SGD \\
\hline
TranSparse & $\| \textbf{W}_{r}^h(\theta_r^h)\textbf{h} + \textbf{r} - \textbf{W}_{r}^t(\theta_r^t)\textbf{t} \|_{\ell_{1/2}}$ ;  $\textbf{W}_{r}^h$, $\textbf{W}_{r}^t$ $\in$ $\mathbb{R}^{d \times k}$; $\theta_r^h$, $\theta_r^t \in \mathbb{R}$ ; $\textbf{r} \in$ $\mathbb{R}^{d}$ & SGD  \\
\hline
\hline
Our STransE & $\| \textbf{W}_{r,1}\textbf{h} + \textbf{r} - \textbf{W}_{r,2}\textbf{t}\|_{\ell_{1/2}}$ ;  $\textbf{W}_{r,1}$, $\textbf{W}_{r,2}$ $\in$ $\mathbb{R}^{k \times k}$; \textbf{r} $\in$ $\mathbb{R}^{k}$ & SGD  \\
\hline
\end{tabular}
}
}
\caption{The score functions $f_r(h, t)$ and the optimization methods
  (Opt.) of several prominent embedding models for KB completion.  
  In all of these the entities $h$ and $t$ are represented
  by  vectors $\textbf{h}$ and $\textbf{t} \in \mathbb{R}^{k}$ respectively.
}
\label{tab:emmethods}
\end{table*}

\emph{Embedding models} for KB completion associate entities and/or
relations with dense feature vectors or matrices.  Such models obtain
state-of-the-art performance
\cite{ICML2011Nickel_438,bordes-2011,Bordes2014SME,NIPS2013_5071,NIPS2013_5028,AAAI148531,guu-miller-liang:2015:EMNLP}
and generalize to large KBs \cite{Denis2015}.
Table~\ref{tab:emmethods} \CHANGEB{summarizes} a number of prominent embedding
models for KB completion.

Let $(h, r, t)$ represent a triple.  In all of the models discussed
here, the head entity $h$ and the tail entity $t$ are represented by
vectors $\textbf{h}$ and $\textbf{t}\in\mathbb{R}^{k}$ respectively.
The \emph{Unstructured} model \cite{Bordes2014SME}
assumes that $\textbf{h} \approx \textbf{t}$. As the Unstructured
model does not take the relationship $r$ into account, it cannot
distinguish different relation types.  The \emph{Structured
  Embedding} (SE) model \cite{bordes-2011} extends the unstructured
model by assuming that $h$ and $t$ are similar only in a
relation-dependent subspace.  It represents each relation $r$ with two
matrices $\textbf{W}_{r,1}$ and $\textbf{W}_{r,2}\in\mathbb{R}^{k\times k}$, which are chosen
so that $\textbf{W}_{r,1}\textbf{h} \approx \textbf{W}_{r,2}\textbf{t}$. The \emph{TransE} model
\cite{NIPS2013_5071} is inspired by models such as Word2Vec
\cite{Mikolov13b} where relationships between words often correspond
to translations in latent feature space.  The TransE model
represents each relation $r$ by a translation
vector \textbf{r} $\in\mathbb{R}^{k}$, which is chosen so that 
$\textbf{h} + \textbf{r} \approx \textbf{t}$.

 The primary contribution of \CHANGEB{this} paper is that two very simple relation-prediction models, SE and TransE, can be combined into a single model, which we call \emph{STransE}.\footnote{Source code: \url{https://github.com/datquocnguyen/STransE}}  
Specifically, we use relation-specific matrices $\textbf{W}_{r,1}$ and
$\textbf{W}_{r,2}$ as in the SE model to identify the
relation-dependent aspects of both $h$ and $t$, and use a vector
$\textbf{r}$ as in the TransE model to describe the relationship
between $h$ and $t$ in this subspace.  Specifically, our new KB completion model STransE 
chooses $\textbf{W}_{r,1}$, $\textbf{W}_{r,2}$ and $\textbf{r}$ so that
$\textbf{W}_{r,1}\textbf{h} + \textbf{r} \approx
\textbf{W}_{r,2}\textbf{t}$\CHANGEB{.  That is, a TransE-style relationship} \CHANGEA{holds in some relation-dependent subspace, and crucially, this subspace may involve very different projections of the head $h$ and tail $t$. So $\textbf{W}_{r,1}$ and $\textbf{W}_{r,2}$ can highlight, suppress, or even change the sign of, relation-specific attributes of $h$ and $t$.  For example, for the ``purchases'' relationship, certain attributes of individuals $h$ (e.g., age, gender, marital status) are presumably strongly correlated with very different attributes of objects $t$ (e.g., sports car, washing machine and the like).}

 As we show below,  STransE 
performs better than the SE and TransE models and other
state-of-the-art link prediction models on two standard link prediction datasets WN18 and FB15k, \CHANGEA{so it can serve as a new baseline for KB completion.  We \CHANGEB{expect} that the STransE \CHANGEB{will also be able to} serve as the basis for extended models that exploit a wider variety of information sources\CHANGEB{,} just as TransE does.}

\section{Our approach}
Let $\mathcal{E}$ denote the set of entities and $\mathcal{R}$ the set
of relation types. For each triple $(h, r, t)$, where $h, t \in
\mathcal{E}$ and $r \in \mathcal{R}$, the STransE model defines a \emph{score
function} $f_r(h, t)$ of its implausibility.
Our goal is to choose $f$ such that the score $f_r(h,t)$ of a plausible triple $(h,r,t)$ is
smaller than the score $f_{r'}(h',t')$ of an implausible triple $(h',r',t')$. We define the STransE score function $f$ as follows:

%
\vspace{-15pt}

\begin{eqnarray*}
f_r(h, t) & = & \| \textbf{W}_{r,1}\textbf{h} + \textbf{r} - \textbf{W}_{r,2}\textbf{t}\|_{\ell_{1/2}} 
\end{eqnarray*}


\noindent using either  the $\ell_1$ or the $\ell_2$-norm \CHANGEB{(the choice is made using validation data; in our experiments we found that the $\ell_1$ norm gave slightly better results)}. 
%
 To learn the vectors and matrices we minimize the following margin-based objective function:
%


\begin{eqnarray*}
\mathcal{L} & = & \sum_{\substack{(h,r,t) \in \mathcal{G} \\ (h',r,t') \in \mathcal{G}'_{(h, r, t)}}} [\gamma + f_r(h, t) - f_r(h', t')]_+
\end{eqnarray*}
%



\noindent where $[x]_+ = \max(0, x)$, $\gamma$ is the margin hyper-parameter, $\mathcal{G}$ is
the training set consisting of correct triples, and
$\mathcal{G}'_{(h, r, t)} = \lbrace (h', r, t) \mid h' \in
\mathcal{E}, (h', r, t) \notin \mathcal{G} \rbrace \cup \lbrace (h, r,
t') \mid t' \in \mathcal{E}, (h, r, t') \notin \mathcal{G} \rbrace $
is the set of incorrect triples generated by corrupting a
correct triple $(h, r, t)\in\mathcal{G}$.  

We use Stochastic
Gradient Descent (SGD) to minimize $\mathcal{L}$, and
impose the following
constraints during training:
$\|\textbf{h}\|_2 \leqslant 1$, $\|\textbf{r}\|_2 \leqslant 1$,
$\|\textbf{t}\|_2 \leqslant 1$, $\| \textbf{W}_{r,1}\textbf{h}\|_2
\leqslant 1$ and $\| \textbf{W}_{r,2}\textbf{t}\|_2 \leqslant 1$.
%
%

\section{Related work}
\label{sec:related}
Table \ref{tab:emmethods} summarizes related embedding models for link
prediction and KB completion.  The models differ in the score
functions $f_r(h, t)$ and the algorithms used to optimize the
margin-based objective function, e.g., SGD, AdaGrad
\cite{Duchi:2011:ASM:1953048.2021068}, AdaDelta
\cite{DBLP:journals/corr/abs-1212-5701} and L-BFGS \cite{Liu1989}.

DISTMULT \cite{yang-etal-2015} is based on a Bilinear model
\cite{ICML2011Nickel_438,Bordes2014SME,NIPS2012_4744}
where each relation is represented by a diagonal rather than a full
matrix. 
The neural tensor network (NTN) model \cite{NIPS2013_5028} uses a
bilinear tensor operator to represent each relation while ProjE \citep{ShiW16a} could be viewed as a simplified version of NTN with diagonal matrices. Similar
quadratic forms are used to model entities and relations in  KG2E \citep{He:2015}, ComplEx \citep{TrouillonWRGB16}, TATEC \citep{Garcia-DuranBUG15} and RSTE \citep{Tay:2017:RST:3018661.3018695}. In addition,  HolE \citep{Nickel:2016:HEK:3016100.3016172} uses  circular correlation---a compositional operator---which could be interpreted as a compression of the tensor product. 

The TransH model \cite{AAAI148531} associates each relation with a
relation-specific hyperplane and uses a projection vector to project
entity vectors onto that hyperplane. TransD
\cite{ji-EtAl:2015:ACL-IJCNLP} and TransR/CTransR \cite{AAAI159571}
extend the TransH model using two projection vectors and a matrix to project entity vectors into
a relation-specific space, respectively. TransD learns a relation-role
specific mapping just as STransE, but represents this mapping by projection
vectors rather than full matrices, as in STransE. The lppTransD  model \citep{yoon-EtAl:2016:N16-1}   extends  TransD to additionally use two projection vectors for representing each relation. 
In fact, our STransE  model and TranSparse   \citep{JiLH016} can be viewed as direct extensions of the TransR model, where head and tail entities are associated with their own projection matrices,
rather than using the same matrix for both, as in TransR and CTransR.

Recently, several authors have shown that relation paths between entities in KBs provide richer information and improve the relationship prediction \citep{lin-EtAl:2015:EMNLP1,garciaduran-bordes-usunier:2015:EMNLP,guu-miller-liang:2015:EMNLP,wang-EtAl:2016:P16-13,feng-EtAl:2016:COLING1,Liu:2016:HRW:2911451.2911509,NIPS2016_6098,wei-zhao-liu:2016:EMNLP2016,toutanova-EtAl:2016:P16-1,NguyenCoNLL2016}. In addition,  \newcite{NickelMTG15} reviews other
approaches for learning from KBs and multi-relational data.

\section{Experiments}
For link prediction evaluation, we conduct experiments and compare the
performance of our STransE model with published results on the
benchmark WN18 and FB15k datasets \cite{NIPS2013_5071}. Information
about these datasets is given in Table \ref{tab:datasets}.

\begin{table}[ht]
\centering
\resizebox{8.25cm}{!}{
\begin{tabular}{l|lllll}
\hline
\bf Dataset & \bf \#E & \bf \#R & \bf \#Train & \bf \#Valid & \bf \#Test \\
\hline
WN18 & 40,943 & 18 & 141,442 & 5,000 & 5,000\\
FB15k & 14,951 & 1,345 & 483,142 & 50,000 & 59,071\\
\hline
\end{tabular}
}
\caption{Statistics of the experimental datasets used in this study
  (and previous works). \#E is the number of entities, \#R is the
  number of relation types, and \#Train, \#Valid and \#Test are the
  numbers of triples in the training, validation and test sets,
  respectively. }
\label{tab:datasets}
\end{table}

\begin{table*}[ht]
\centering
\resizebox{16cm}{!}{
\begin{tabular}{l|lll|lll|lll|lll}
\hline
\multirow{3}{*}{\bf Method}& \multicolumn{6}{|c}{\textbf{Raw}}& \multicolumn{6}{|c}{\textbf{Filtered}} \\
\cline{2-13}
& \multicolumn{3}{|c}{\bf WN18} & \multicolumn{3}{|c}{\bf FB15k} & \multicolumn{3}{|c}{\bf WN18} & \multicolumn{3}{|c}{\bf FB15k}\\
\cline{2-13}
\cline{2-13}
&   MR & H10 & MRR &   MR & H10 & MRR &   MR & H10 & MRR &   MR & H10 & MRR \\
\hline
SE \citep{bordes-2011} & 1011 & 68.5 & - & 273& 28.8 & - & 985 &80.5 & - &  162  & 39.8& - \\
Unstructured \citep{Bordes2014SME}  & 315  & 35.3 & - &1074 & 4.5 & - &  304 & 38.2  &-&   979 &  6.3& - \\
 TransE \citep{NIPS2013_5071}  & 263 & 75.4 & - & 243 & 34.9 & - & 251 & 89.2 & -  &  125 & 47.1 & - \\
TransH \citep{AAAI148531}  & 401 & 73.0 & - & 212& 45.7 & -&  303 & 86.7 & - &  87 &  64.4&- \\
TransR \citep{AAAI159571}  & 238& 79.8 & -  & 198&48.2 & -  &  225 &92.0 & - &   77 &   68.7&- \\
CTransR \citep{AAAI159571}  & 231& 79.4 & - & 199& 48.4& - & 218 &  92.3& -  &   75  & 70.2 &  -\\
KG2E \citep{He:2015}  &342& 80.2  & - & \textbf{174} & 48.9& -   &  331 & 92.8 & - &  59 & 74.0& -  \\
TransD \citep{ji-EtAl:2015:ACL-IJCNLP}  & 224 & 79.6 & - & 194 & 53.4 & - &  212 &  92.2 & - &   91 & {77.3}&  -  \\
lppTransD \citep{yoon-EtAl:2016:N16-1} & 283 & 80.5& -  & 195 & 53.0 & - & 270 & 94.3 & - & 78  & 78.7 & -\\
TranSparse \citep{JiLH016} & 223 & 80.1& -  & {187} & \textbf{53.5} & -  & 211  & 93.2 & - & 82  & 79.5& -\\

TATEC \citep{Garcia-DuranBUG15}  & - & -  &- & -&- & -& - &  - & - &   \textbf{58} & 76.7  & -  \\
NTN \citep{NIPS2013_5028}  &- & -  & -& -& -&- & - & {66.1}  & 0.53& - & 41.4& 0.25  \\
DISTMULT \citep{yang-etal-2015}  & -& - & -& -&- &- & - & {{94.2}} & 0.83 & -  & 57.7 & 0.35\\
HolE \citep{Nickel:2016:HEK:3016100.3016172} & -&- &\textbf{0.616} &- &- &0.232& -& \textbf{94.9} & \textbf{0.938} & - & 73.9 & 0.524\\ 
\hline
Our STransE   & \textbf{217} & \textbf{80.9} & 0.469 & 219 & 51.6 & \textbf{0.252} &  \textbf{206}  & {93.4}& 0.657 & 69  & \textbf{79.7}& \textbf{0.543}\\
\hline
\hline
\textsc{r}TransE \citep{garciaduran-bordes-usunier:2015:EMNLP}  & -&  - &- & -&- &- &  -& - & - & {\textbf{50}} & 76.2  & - \\
PTransE \citep{lin-EtAl:2015:EMNLP1}  & -& - & -& 207& 51.4& -  & - & -  & - & {58} & {84.6}& - \\
GAKE \citep{feng-EtAl:2016:COLING1} & - & - & - & 228  &44.5 &-& - & - & - & 119 & 64.8& - \\ 
Gaifman \citep{NIPS2016_6098} & - & - & - & - & - &- & 352  & 93.9 & -& 75 & 84.2 & - \\ 
Hiri \citep{Liu:2016:HRW:2911451.2911509} & - & - & - & - & - & - & -  & 90.8 & 0.691& -  & 70.3& 0.603 \\
\hline
NLFeat \citep{toutanova-chen:2015:CVSC}  &- & - &- & -&- &- & -   & \textbf{94.3} & \textbf{0.940}& -  & \textbf{87.0}& \textbf{0.822} \\
TEKE\_H  \citep{DBLP:conf/ijcai/WangL16} & \textbf{127}  & 80.3 & -& 212  & 51.2 & -&  \textbf{114}  & 92.9 & -& 108 & 73.0& -  \\ 
SSP \citep{0005HZ16} & 168 & \textbf{81.2} & - & \textbf{163} & \textbf{57.2} & - & 156 & 93.2 & - & 82 & 79.0 & -\\ 
\hline
\end{tabular}
}
\caption{Link prediction results. MR, H10 and MRR denote evaluation metrics of mean rank, Hits@10 (in \%) and mean reciprocal rank, respectively. ``NLFeat'' abbreviates Node+LinkFeat. The results for NTN {\protect\citep{NIPS2013_5028}} listed in this table are taken from {\protect\citet{yang-etal-2015}} since  NTN was  originally evaluated on different datasets. }
\label{02tab:linkprediction}
\end{table*}

\subsection{Task and evaluation protocol}
The link prediction task
\cite{bordes-2011,Bordes2014SME,NIPS2013_5071} \CHANGEB{predicts} the head or tail entity given the relation type and
the other \CHANGEB{entity}, i.e. predicting $h$ given $(?, r, t)$ or predicting
$t$ given $(h, r, ?)$ where $?$ denotes the missing element. The results are evaluated using the ranking  induced by the score function $f_r(h,t)$ on test triples.

For each test triple $(h, r, t)$, we corrupted it by replacing either $h$ or $t$ by each of {the possible} 
entities in turn, and then rank these candidates in ascending order of
their implausibility value computed by the score function. This is called as the ``Raw'' setting
protocol. For the ``Filtered'' setting protocol described
in  \citet{NIPS2013_5071}, 
{we  \textit{removed} any corrupted triples}
that appear in the knowledge base,
to avoid cases where a correct corrupted triple might be  ranked higher than the
test triple.  The ``Filtered'' setting thus provides
a clearer view on the ranking performance. 
Following \citet{NIPS2013_5071}, we report the mean rank and the 
Hits@10  (i.e., the proportion of test triples in which
the target entity was ranked in the top 10 predictions) for each model. In addition,  we  report the
mean reciprocal rank, which is commonly
used in information retrieval. 
In both ``Raw'' and
``Filtered'' settings, lower mean
rank, higher mean reciprocal rank or higher Hits@10 indicates better link prediction performance.

Following TransR  \citep{AAAI159571}, TransD
\citep{ji-EtAl:2015:ACL-IJCNLP}, 
\textsc{r}TransE \citep{garciaduran-bordes-usunier:2015:EMNLP},
PTransE \citep{lin-EtAl:2015:EMNLP1}, TATEC \citep{Garcia-DuranBUG15} and TranSparse \citep{JiLH016}, 
we used the entity and relation
vectors produced by TransE \cite{NIPS2013_5071} to initialize the entity
and relation vectors in STransE\CHANGEB{, and} we initialized the relation matrices
with identity matrices.  
We applied the ``Bernoulli'' trick used also in previous work for generating head or tail entities when sampling incorrect triples \citep{AAAI148531,AAAI159571,He:2015,ji-EtAl:2015:ACL-IJCNLP,lin-EtAl:2015:EMNLP1,yoon-EtAl:2016:N16-1,JiLH016}. 
We ran SGD for 2,000 epochs to estimate the
model parameters.  Following \newcite{NIPS2013_5071} we used a grid
search on \CHANGEB{validation set to choose either the $l_1$ or $l_2$ norm in the score function $f$, as well as to} set the SGD learning rate
$\lambda\in\lbrace 0.0001, 0.0005, 0.001, 0.005, 0.01 \rbrace$, the
margin hyper-parameter $\gamma\in\lbrace 1, 3, 5 \rbrace$ and the vector size  $k\in\lbrace 50, 100 \rbrace$.  The lowest filtered mean rank on the
\CHANGEB{validation} set was obtained when \CHANGEB{using the $l_1$ norm in $f$ on both
WN18 and FB15k, and when} $\lambda = 0.0005, \gamma = 5,
\text{ and } k = 50$ for WN18, and $\lambda = 0.0001, \gamma = 1,
\text{ and } k = 100$ for FB15k. 

\subsection{Main results}
Table \ref{02tab:linkprediction} compares the link prediction results of
our STransE model with results reported in prior work, using the same
experimental setup. The first 15 rows report the performance of
the models that do not exploit information about alternative paths between
head and tail entities. The next 5 rows report results of the  models that exploit information about relation paths. The last 3
rows present  results for the models  which make  use of
textual mentions derived from a large external corpus.

It is clear that the models with the additional external corpus
information obtained best results. In future work we plan to extend
the STransE model to incorporate such additional information. Table 
\ref{02tab:linkprediction}  also shows that the models   employing path information generally achieve better results
than models that do not use such information. In terms of models not
exploiting path information or external information, the STransE model
produces the highest filtered mean rank  on WN18 and  the highest
filtered Hits@10 and mean reciprocal rank on FB15k. Compared to the closely related models SE, TransE, TransR,
CTransR, TransD and TranSparse,  our STransE model does better than these models on both WN18
and FB15k.

Following \newcite{NIPS2013_5071}, Table \ref{tab:headtail} analyzes  
Hits@10 results on FB15k with respect to the relation categories
defined as follows: for each relation type $r$, we computed the
averaged number $a_h$ of heads $h$ for a pair $(r, t)$ and the averaged
number $a_t$ of tails $t$ for a pair $(h, r)$. If $a_h < 1.5$ and $a_t
< 1.5$, then $r$ is labeled \textbf{1-1}. If $a_h \geq 1.5$ and $a_t
< 1.5$, then $r$ is labeled \textbf{M-1}. If $a_h < 1.5$ and $a_t
\geq 1.5$, then $r$ is labeled as \textbf{1-M}. If $a_h \geq 1.5$
and $a_t \geq 1.5$, then $r$ is labeled as \textbf{M-M}. 1.4\%,
8.9\%, 14.6\% and 75.1\% of the test triples belong to a relation type
classified as  \textbf{1-1}, \textbf{1-M}, \textbf{M-1} and \textbf{M-M},
respectively.

\begin{table}[ht]
\centering
\resizebox{8cm}{!}{
\setlength{\tabcolsep}{0.25em}
\begin{tabular}{l|llll|llll}
\hline
\multirow{2}{*}{\bf Method}& \multicolumn{4}{|c}{\bf Predicting head $h$} & \multicolumn{4}{|c}{\bf Predicting tail  $t$}\\
\cline{2-9}
& 1-1 & 1-M & M-1 & M-M & 1-1 & 1-M & M-1 & M-M \\
\hline
SE  & 35.6 & 62.6 & 17.2 & 37.5 & 34.9 & 14.6 & 68.3 & 41.3 \\
Unstr. & 34.5 & 2.5 & 6.1 & 6.6  & 34.3 & 4.2 & 1.9 & 6.6\\
 TransE  & 43.7 & 65.7 & 18.2 & 47.2 & 43.7 & 19.7 & 66.7 & 50.0  \\
TransH  & 66.8 & 87.6 & 28.7 & 64.5 & 65.5 & 39.8 & 83.3 & 67.2 \\
TransR  & 78.8 & 89.2 & 34.1 & 69.2 & 79.2 & 37.4 & 90.4 & 72.1 \\
CTransR   & 81.5 & 89.0 & 34.7 & 71.2  & 80.8 & 38.6 & 90.1 & 73.8 \\
KG2E   & \textbf{92.3} & {94.6} & \textbf{66.0} & 69.6 &  \textbf{92.6} & \textbf{67.9} & {94.4} & 73.4  \\
TATEC    & 79.3 & 93.2 & 42.3 & 77.2 & 78.5 & 51.5 & 92.7 & 80.7 \\
TransD   & 86.1 & \textbf{95.5} & 39.8 & 78.5 & 85.4 & 50.6 & {94.4} & 81.2  \\
lppTransD & 86.0 & 94.2 & 54.4 & \textbf{82.2} & 79.7 & 43.2 & \textbf{95.3} & 79.7\\
TranSparse & 86.8 & \textbf{95.5} & 44.3 & {80.9} & 86.6 & 56.6 & 94.4 & \textbf{83.3}\\
\hline
STransE & {82.8} & {94.2} & {50.4} & {80.1} & {82.4} & {56.9} & {93.4} & {83.1}\\ 

\hline
\end{tabular}
}
\caption{Hits@10 (in \%) by the relation category on  FB15k. ``Unstr.'' abbreviates Unstructured.}
\label{tab:headtail}
\end{table}

Table \ref{tab:headtail} shows that in comparison to
prior models not using path information, STransE obtains the second highest
Hits@10 result for \textbf{M-M} relation category at $(80.1\% + 83.1\%) / 2 = 81.6\%$  which is  0.5\% smaller than the Hits@10 result of TranSparse  for \textbf{M-M}. However, STransE obtains 2.5\% higher Hits@10 result than TranSparse for \textbf{M-1}. In
addition, STransE also performs better than TransD for \textbf{1-M} and
\textbf{M-1} relation categories. We believe the improved performance of
the STransE model is due to its use of full matrices\CHANGEB{,} rather than \CHANGEB{just}
projection vectors as in TransD. This permits STransE to model diverse
and complex relation categories (such as \textbf{1-M}, \textbf{M-1} and
especially \textbf{M-M}) better than TransD \CHANGEA{and other \CHANGEB{similiar} models}. However, STransE is not
as good as TransD for the \textbf{1-1} \CHANGEB{relations}. 
\CHANGEB{Perhaps the extra parameters in STransE hurt performance in this case
(note that 1-1 relations are relatively rare, so STransE does better overall).}

\section{Conclusion and future work}


This paper presented a new embedding model for link prediction and KB completion. \CHANGEA{Our STransE combines insights \CHANGEB{from} several simpler embedding models, specifically \CHANGEB{the} Structured Embedding \CHANGEB{model} \cite{bordes-2011} and \CHANGEB{the} TransE \CHANGEB{model} \cite{NIPS2013_5071}}\CHANGEB{,} by using a low-dimensional vector and two \CHANGEB{projection} matrices to represent each relation. STransE, \CHANGEA{while being conceptually simple}, produces highly competitive results on standard link prediction evaluations, \CHANGEA{and scores better than the embedding-based models it builds on. Thus it is a suitable candidate for serving as future baseline for more complex models in the link prediction task.}

 In future work we plan to extend STransE to exploit relation path information in knowledge bases, in a manner similar to \newcite{lin-EtAl:2015:EMNLP1}, 
\newcite{guu-miller-liang:2015:EMNLP} or \newcite{NguyenCoNLL2016}.

\section*{Acknowledgments}  
This research was supported by a Google award through the Natural 
Language Understanding Focused Program, and under the Australian 
Research Council's {\em Discovery Projects} funding scheme (project 
number DP160102156).

NICTA is funded by the Australian Government through the Department of Communications and the Australian Research Council through the ICT Centre of  Excellence Program. The first author is supported by an International Postgraduate Research Scholarship and a NICTA NRPA Top-Up Scholarship. 

\bibliographystyle{naaclhlt2016}
\bibliography{REFs}

\end{document}